\documentclass[preprint,review,3p,times]{elsarticle}

\usepackage{lineno}
\usepackage{booktabs}
\usepackage{textcomp}
\usepackage{mathrsfs}
\usepackage[english]{babel}
\usepackage{graphicx}
\usepackage{nomencl}
\makenomenclature
\biboptions{sort&compress}
\usepackage{relsize}
\usepackage{amsmath}
\usepackage{amssymb}
\usepackage[linesnumbered,boxruled,commentsnumbered]{algorithm2e}

\SetCommentSty{mycommfont}
\usepackage{lineno}
\usepackage{amsfonts}
\usepackage{float}
\usepackage{longtable}
\usepackage{caption}
\usepackage{appendix}
\usepackage{enumerate}
\usepackage{esdiff}
\usepackage{breqn}
\usepackage[colorlinks]{hyperref}
\usepackage{footnote}
\usepackage{fixfoot}
\usepackage{ragged2e}
\usepackage{tabularx, booktabs, threeparttable, caption}
\usepackage{natbib}
\usepackage{multirow}
\usepackage[export]{adjustbox}
\usepackage[most]{tcolorbox}
\usepackage{lscape}
\usepackage{rotating}
\usepackage{longtable}
\usepackage{seqsplit}
\let\oldseqsplit\seqsplit
\renewcommand{\seqsplit}{
	\expandafter\oldseqsplit\expandafter}%
\usepackage{array}
\newcolumntype{C}{ >{\centering\arraybackslash} m{1.0cm} }
\newcolumntype{Q}{ >{\centering\arraybackslash} m{3 cmcm} }
\newcolumntype{M}[1]{>{\centering\arraybackslash}m{#1}}
\newcolumntype{P}[1]{>{\centering\arraybackslash}p{#1}}

\usepackage{mathrsfs}
\usepackage{breakurl}

\newcounter{expcnt}
\newlength\mylength

\begin{document}

	\begin{frontmatter}

		\title{A Review on the Applications of Transformer-based language models for Nucleotide Sequence Analysis}

		\author[1,2, *]{Nimisha Ghosh\corref{mycorrespondingauthor}}%
             \author[2,*]{Daniele Santoni}
		\author[3]{Indrajit Saha}%
            \author[2]{Giovanni Felici}

		\address[1]{Department of Computer Science and Information Technology, Institute of Technical Education and Research, \\Siksha `O' Anusandhan (Deemed to be University), Bhubaneswar, Odisha, India}
    \address[2]{Institute for System Analysis and Computer Science “Antonio Ruberti”, National Research Council of Italy, Rome, Italy}
		\address[3]{Department of Computer Science and Engineering, \\National Institute of Technical Teachers' Training and Research, Kolkata, West Bengal, India}

		\address[*]{Equally contributed}
		\cortext[mycorrespondingauthor]{Corresponding author: nimishaghosh@soa.ac.in}
		\begin{abstract}
 In recent times, Transformer-based language models are making quite an impact in the field of natural language processing. As relevant parallels can be drawn between biological sequences and natural languages, the models used in NLP can be easily extended and adapted for various applications in bioinformatics. In this regard, this paper introduces the major developments of Transformer-based models in the recent past in the context of nucleotide sequences. We have reviewed and analysed a large number of application-based papers on this subject, giving evidence of the main characterizing features and to different approaches that may be adopted to customize such powerful computational machines. We have also provided a structured description of the functioning of Transformers, that may enable even first time users to grab the essence of such complex architectures. We believe this review will help the scientific community in understanding the various applications of Transformer-based language models to nucleotide sequences. This work will motivate the readers to build on these methodologies to tackle also various other problems in the field of bioinformatics.
\end{abstract}

	\end{frontmatter}

\section{Introduction}
Starting from the 1940s, scientists began to study the amino acid composition of proteins in order to characterize tissues and species by their amino acid frequencies \cite{BEACH1943}. Insulin was the first protein to be sequenced in 1951 and 1952 (chains of bovine insulin B and A, respectively) by Sanger \cite{Sanger49} that led to him winning the Nobel in chemistry (1958). In the following years, as new technologies were developed, many other
sequences became available thereby opening new frontiers in Biology and Chemistry. In fact, in the 70s a consistent number of sequences was made available, and several new issues related to protein sequences arose. 
Many works show that classification methods are able to separate proteins into different families \cite{Ferran1991, Orengo1993,blekas2005, Exarchos2006,Kocsor2005}) based exclusively on their sequences.
Also at genomic level, as a significant number of fully sequenced genomes was made available, scientists started to study and compare genome features in terms of similarity, complexity, information content and statistical properties. In the late 90s, the first whole-genome studies provided insights on genome composition in terms of subsequences or \textit{k}-mers \cite{karlin1997, Karlin97}.
Recently, Natural Language Processing (NLP) approaches have been showing a significant impact in bioinformatics ranging from DNA/RNA sequence analysis to computational biology \cite{IUCHI2021, Zeng2018}. Since biological sequences can be seen as \textit{words} on given alphabets (the four nucleotides for genomic sequences) or as texts (where words are \textit{k}-mers) an NLP approach seems to be particularly suitable and effective to investigate and extract information in this context. 

For more than two decades now, the abilities of neural networks in NLP tasks have been exploited, and series of word embedding technologies have been used for unique representation of text \cite{Blacoe2012, Turian2010}.

As an example, Word2Vec \cite{Mikolov2013a, Mikolov2013b} can be considered as a representative model which uses shallow neural network for obtaining vector representation of words. Word2vec utilises either Continuous Skip-Gram (CSG) or Continuous Bag-of-Words (CBOW). Given a current word, CSG predicts the surrounding words while CBOW predicts the current word based on the context. The evolution of language models in NLP can be accredited to the rapid development of deep learning technologies. Conventional RNN-based models such as Bidirectional Recurrent Neural Network (Bi-RNN) \cite{Schuster1997}, Long-Short Term Memory (LSTM) \cite{Hochreiter1997} and Gated Recurrent Unit (GRU) \cite{cho2014} attempt to encode the entire sequence into a finite length vector. 
Such deep learning models have been extensively used in bioinformatics, e.g., for the prediction of Transcription Factor Binding Sites \cite{Zeng2020review}. However, such models suffer from vanishing gradient (gradients become too small during backpropagation) as well as low-efficiency problems. Subsequently, Transformer \cite{NIPS2017} models came into existence which did not have any RNN-based network structures and was dependent on multi-head attention mechanism (described later in Section 2) which focuses on computing similarity scores between words in a sentence. The advantage of Transformers is that they are not dependent on past hidden states to understand the dependency on previous words. To avoid vanishing gradient problem and performance deterioration caused by long-term dependency, Transformers process a sentence as a whole to enable parallel computation.  Considering the aforementioned advantages of parallel computation and multi-head attention, this review article  focuses on the use of Transformer-based language models in bioinformatics, in particular their applications in processing nucleotide sequences. 

Transformer-based language models can be categorised into either trained from scratch or pre-trained models. Models which are trained from scratch are based on the direct training of all model parameters from the initial stages using datasets specific to a task, and takes many iterations to completely converge. One such example is Transformer-XL \cite{Dai2019} which uses segment-level recurrent mechanism and a novel position encoding scheme to model long text. Sparse Transformers \cite{zhao2019} by Zhao et al. have introduced a variation in architecture of traditional Transformers to train deeper networks as well as have recomputed attention matrices to save memory. To improve the efficiency of Transformers, Kitaev et al. \cite{Kitaev2020} have proposed Reformer which uses locality-sensitive hashing and reversible residual layers. Although self-attention is a very powerful component in Transformers, its computational and memory requirements increase quadratically with the length of a sequence. Thus, it becomes either infeasible or very expensive to process long sequences. To mitigate this problem, Beltagy et al. \cite{beltagy2020} have proposed Longformer with a self-attention mechanism that scales linearly with
the sequence length. 

Transformer-based pre-trained models, on the other hand, are trained using a vast amount of unlabeled data before being finetuned for particular tasks. Pre-training increases the convergence rate of the target tasks, learns generic knowledge from unlabeled input, and typically has better generalization than training parameters from scratch \cite{HAN2021ai}. Models are then finetuned on specific tasks considering labeled data and subsequently a classification layer is built on top for the prediction of corresponding tasks (the concepts of pre-training and finetuning are provided in details in Section 2). In this regard, GPT-X \cite{radford2018, radford2019, Brown2020} is a first of its kind finetuned Transformer model. BERT \cite{devlin2019} is yet another finetuned model which uses bi-directional Transformers and mask mechanism. Dynamic masking is used by RoBERTa \cite{liu2019roberta} and it has shown better performance than BERT. Based on Transformer-XL architecture, XLNet \cite{Yang2019xlnet} has permutation language modelling that has improved training method. ERNIE \cite{zhang2019ernie} as proposed by Zhang et al. has a continual learning mechanism consisting of two parts; continual construction of pre-training tasks and incremental multi-task learning. Raffel et al. \cite{Raffel2020t5} have proposed T5 which is a unified framework to convert all text-based language problems into a text-to-text format. In this regard, T5 has achieved competent results on many areas which encompass question answering, summarisation, text classification etc. To lower memory consumption and increase the training speed of BERT, two parameter reduction techniques such as cross-layer parameter sharing and inter-sentence coherence loss have been used in ALBERT \cite{lan2020albert}.

These aforementioned breakthrough works have prompted many researchers to take up the challenge of applying such methodologies in computational biology. 
Though there have been some significant reviews \cite{Zeng2020review, Zhang2023_1, Martinek2022, IUCHI2021}, the applications of Transformer-based language models for processing nucleotide sequences are in clearly dearth. In this regard, the basics of Transformer framework are put forth which is followed by the review of applications of Transformers for nucleotide sequences. We have divided the considered studies into subsections according to the biological context they are referred to.   
We hope that this review will benefit the research community in understanding the applications of Transformer-based models in bioinformatics.

\section{The Essence of Transformers}

Moving forward with our description, we continue with the example of natural language processing: we assume to have a corpus made of a number of {\it sentences}, composed of {\it words}; such words belong to a finite {\it vocabulary}. Such sentences may be in a special relationship, i.e., one sentence may be connected to another because it is the answer to a question posed by the first sentence, or its translation; when such relations are present, we will consider the first sentence to be the input and the second to be the output. Changing the alphabet to nucleotides and sentences to DNA sequences respectively would, with a sufficient level of approximation, allow to use Transformers in bioinformatics. For the ease of understanding and readability, words and tokens are used equivalently to describe the workings of Transformers.

Transformers \cite{NIPS2017} are the most recent innovation in the field of AI and deep learning, and have provided a major breakthrough in natural language processing (NLP). The main feature of Transformers, when applied to NLP, is their ability to understand the intrinsic relations among words within a sentence and those of sentences within a text. Relying on the many similarities between natural languages and biological sequences, Transformer based models have been adapted in bioinformatics with a remarkable degree of success. 
In NLP environment, Transformers can be used to generate text in response to a question, to translate text among languages, to classify pieces of text into classes, and to summarize text.
Transformers work as any other machine learning method: they are composed of a number of mathematical formulas defined by parameters, are submitted a number of examples, and adjust the parameters of the mathematical formulas in such a a way that the output of the formulas is as close as possible to a expected value. Such value is typically the average of a certain error measure over all training samples. 
One of the main tools adopted within Transformers are Neural Networks. The methods adopted to adjust the parameters of the Transformers in the desired directions may be of different types, but the most commonly used ones are the classic {\it backpropagation} method combined with {\it Gradient Descent} and variants. 
Another important feature of Transformers is that they are {\it very large} - as a matter of fact, their most visible application are the Large Language Models (LLM) of ChatGPT, BERT, etc. The main components of Transformers (Encoder and Decoder) are stacked multiple times; the amount of training data, and the cost of training (e.g. finding the best value of the many millions of parameters they contain) is terrific. Their success may also be attributed to the large size of their model and the capability of training such model with large data.

\subsection {PreTraining and FineTuning}

As mentioned earlier, Transformers are a learning mechanism; their learning proceeds with trial and error. The many parameters that define the system are initialized at random, then an input-output pair is passed through it, and an output is predicted; if it is too distant from the actual output, backpropagation is used to modify the parameters (weights) of the system to try and reduce such an error. This is done iteratively for all the training samples, several times, until the error is small enough. But how does this process declinate for the Transformer and language approach? The training is mainly of two types, according to the objective of the application: {\it pre-training} and {\it fine tuning}. The easiest way to look at it, is that pre-training is unsupervised learning, while finetuning is supervised learning. In pre-training, some words in the input are {\it masked}; the system then aims to predict a masked word according to the words that precede them; this way, a prediction error may arise and such an error is backpropagated to adjust the weights. In pre-training, one may assume that the network, through adjusting its weight to reduce errors on masked words, learns the structure of the language, the similarity among words, their relations and the like. Once the system has been pre-trained, finetuning for a specific task may take place: according to a series of input-output training pairs, the weights in the Transformers are further adjusted to generate an output when provided with an input. This is the case of question answering, translation, or text classification: for example, the system may be provided with a large number of sentences in one language and their translation in another language, and learn to translate.

\subsection{Encoding and Decoding}

\begin{figure}[H]
	\centerline{
		\includegraphics[height=2.2in,width=5.0in]{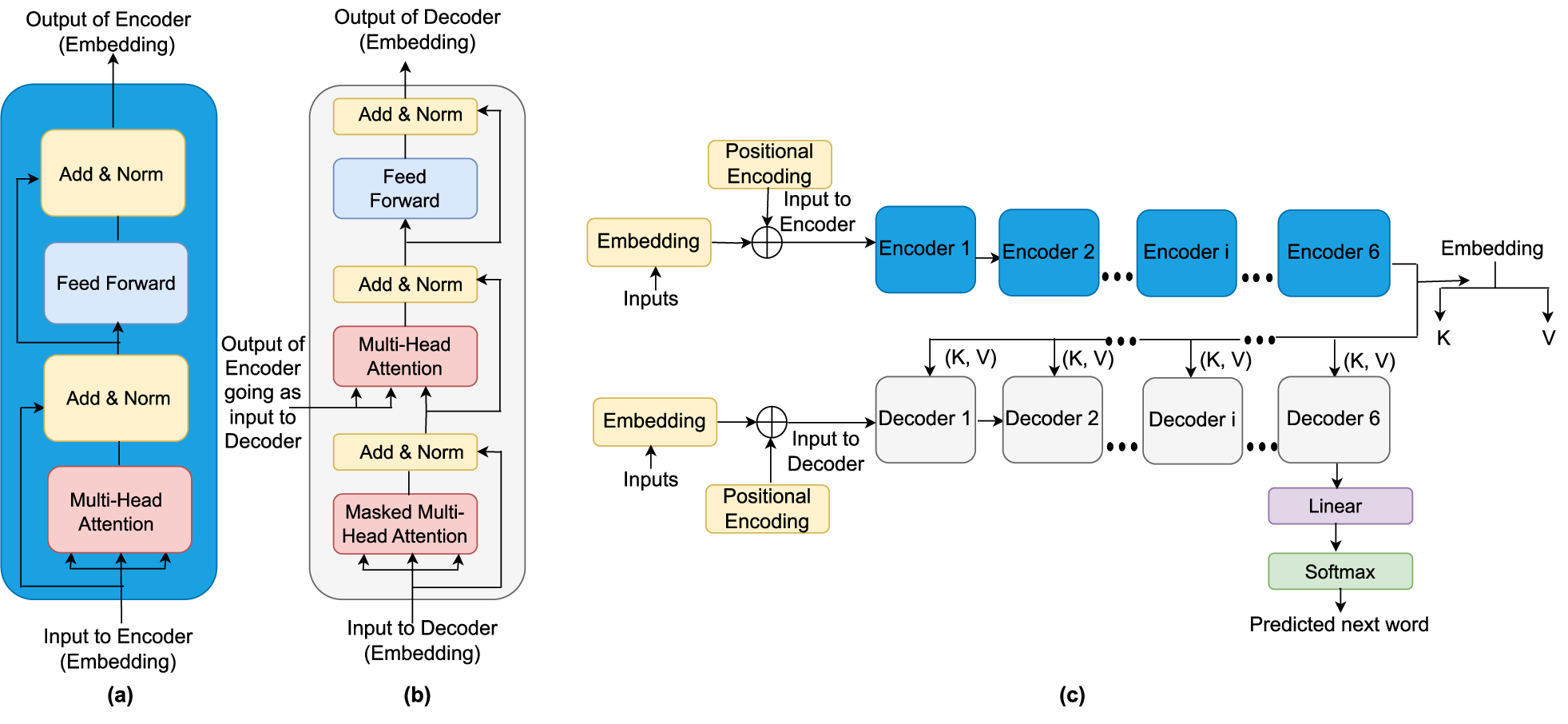}}
	\caption{The figure depicts the basic architecture of a \textbf{Transformer} and its main components: in Pane (a) is a schematic view of an \textbf{Encoder}; in pane (b) that of a \textbf{Decoder}, and in pane (c) the overall combination of stacked Encoders, Decoders, and additional components. To explain the general mechanism, we can consider an example of a training set of \textit{question}-\textit{answer} pairs. The Encoder in (a) takes as input an embedding, i.e., a matrix of fixed dimension, and outputs another  matrix that is the results of the transformation of the input embedding; in Encoder 1 of the stack, the input matrix is created by combining word embedding  and positional encoding of the input question, while for the rest of the Encoders ($i$ in $2$,...,$6$) the input is the embedding as generated by the Encoder ($i-1$). The embedded output of the last Encoder is copied to generate the pair of Key and Value matrices $(K,V)$ and provided as part of the input to all the Decoders (pane (c)). As the first Encoder is in charge of processing the \textit{question} part, the first Decoder plays the same role for the \textit{answer} part. The answer is embedded and combined with positional encoding, and then transformed along the stack of Decoders. The output of the last Decoder is subject to a linear transformation and a \textit{softmax} function. The linear layer transformation has as many output nodes as the length of the vocabulary, and the \textit{softmax} function selects the word in the vocabulary with highest likelihood as the predicted next word. Both Encoders and Decoders contain feed-forward Neural Networks, whose weights are tuned in the training phase based on prediction errors. Moreover, weights are present in the different multi-head attention mechanisms, whose values are tuned as well by the backpropagation process.}
  
	\label{fig1}
\end{figure}

One common way of describing a Transformer is as something composed of an {\bf Encoder} and a {\bf Decoder}. These are two paired machines: the Encoder takes a sentence and embeds it into a set of numbers; the Decoder has the purpose of generating text, one word at a time, based on the embedding of an input sentence. Fig. \ref{fig1}(a) and \ref{fig1}(b) respectively show the different components of Encoders and Decoders. As shown in Fig. \ref{fig1}(c), there are multiple blocks of Encoder and Decoder in a Transformer.

We start from describing the main steps of the Encoder when one sentence is presented to it:

\begin{enumerate}
\item  each word in the sentence is associated with an \textbf{id}; the same word will receive the same \textbf{id} also when it appears in other sentences. If the number of words in the sentence exceeds a maximum value $m$, the sentence will be divided into chunks, and each chunk is considered as a separate sentence; if the length is below $m$, the sequence of \textbf{id}s associated to the words of the sentence will be filled up with padding.

\item the sentence is now represented by a vector of length $m$. Such vector is then expanded into a matrix of $n$ rows of dimension $m$. The parameter $n$ represents the number of additional features that are attached to each word. Such expansion may be viewed as an enrichment of the role of each word, that is allowed to be described with an additional set of $n$ numbers. The dimension of such enrichment is a design decision; the values of these $n$ numbers to be associated with each word are initialized randomly and refined in the training process.

\item It is important here to note that the words of a sentence are fed in parallel to an Encoder, thus the
order in which the words appear in the sentence is not directly taken into account. This is taken care of by \textit{positional encoding}. Furthermore, the main essence of Transformers lies in introducing \textit{self-attention mechanism}, where the relative position of the words in a sentence are explored to identify the relationship among such words. We skip the details for brevity. Interested readers are requested to look into \cite{NIPS2017} for further clarification.


\item At this point, the matrix representing the sentence, also referred to as $Q$ {\it for Query}, is processed via a number of transformations based on different types of neural networks, that have the purpose of extracting the relevant information from each matrix, aligning their dimensions, and combining them with typical neural network activation function. Here, two additional matrices of that same dimensions as $Q$:$K$ and $V$ (for {\it Key} and {\it Value}) are introduced; the three matrices $Q,K,V$ are each passed through a linear network, that combines their values with additional weights. Subsequently, these matrices are combined together with a non-linear network that uses a \textit{RELU} activation and a \textit{softmax} function. The main purpose of $K$ and $V$ is to add additional degrees of freedom to the system, with additional parameters that can be modified in the training;

\item At the end of the encoding process, which encompasses several stacked layers of Encoders, the output is an embedding that, through the many transformation undertaken by the network layers, incorporates the original sentence, the relation among its words, and the weightage associated with the words of the sentences. Such Embedding is then copied into a $(K,V)$ matrix pair and provided as one of the inputs to a stacked layer of Decoders.

\end{enumerate}

Two important facts have to be noted here. First, the Encoder produces the ideal multidimensional numeric features attached to each word of the sentence. The sentence, provided as input, is encoded into a much higher dimensional space where the relation among the words and their position is represented by the values in the output $K$ and $V$. Second, the encoding step 4 is repeated several times, and their output packed together: such mechanism is called {\it multi-head attention} and results into an additional injection of degrees of freedom to the system. It is to be noted that \textit{multi-head attention} is the concatenation of several heads which follow the \textit{self-attention} mechanism as mentioned earlier. The final product of the Encoder is the sequential passes of the steps as described above (typically, in NLP applications, 6 layers are used \cite{NIPS2017}). This is also depicted in Fig. \ref{fig1}. 

The functioning of the {\bf Decoder} has many parts in common with the Encoder, yet there are some major differences. In a nutshell, the Decoder processes the output component of the input-output pairs of the training data; encodes such information and produces its own embedding using a multi-head attention scheme similar to the Encoder, but:

\begin{enumerate}

\item{} once the output has been encoded, it is combined with the  $K$ and $V$ matrices coming from the Encoder. This is the step where Transformers learn the relation between the input of the training (e.g., the question) and the output (e.g., the answer to the question); considering the question-answer example, the Decoder combines, through similar network training mechanisms, the processed embedding of the question and the processed embedding of the answer; 

\item{} at the end of the process a linear layer has a number of output neurons equal to the size of the vocabulary; such network uses a \textit{softmax} function to produce likelihood for each term in the vocabulary. Then, the term with the highest likelihood is the output of the Decoder i.e. the predicted next word;

\item {} The Decoder produces the output one word at a time; in the training phase, knowing the correct word, it can determine the error that then instructs the backpropagation and the correction of the weights.

\end{enumerate}

The steps as mentioned above can be visualised in Fig. \ref{fig1}(c). To fully appreciate the subtleties of the Encoder-Decoder mechanism, one should be aware that Encoder and Decoder are trained simultaneously, both in pre-training and in finetuning. The weights obtained in pre-training are based on masked words and then refined in finetuning, when input-output training pairs are presented to the system. 

A simplified model of Transformer may consider only an extended version of the Encoder, that omits the decoding part; this is the case when there is no need to generate sentences, but simply to produce a classification of the input sentence into a finite number of classes. 
In such a case, the Encoder itself is equipped with a masking mechanisms and with a final neural network with a number of output neurons equal to the number of classes that define the classification problem (remember that, in the case of the standard Decoder, the final neural network layer has a number of output neurons equal to the size of the vocabulary).

Far from aiming at providing a complete description of what Transformers are and how they work, we shed light on the main mechanisms that are behind their effectiveness and on why they appear to work well also for the languages as spoken by nature encompassing nucleotide sequences. The next  section discusses some important models based on Transformers applied on nucleotide sequences.

\section{Applications of Transformer-based language models for Nucleotide sequences}

In this section, a summary of different applications of Transformer-based language models for nucleotide sequences is provided along with their main advantages. 
As reported in Table~\ref{tab:1} the considered studies are partitioned into six areas according to the biological context they refer to: Promoter and Enhancer, Methylation, Reads, Binding and Miscellaneous. 

\subsection{Promoter and Enhancer}
\begin{figure}[H]
				
				\centerline{
					\includegraphics[height=2.0in,width=1.2in]{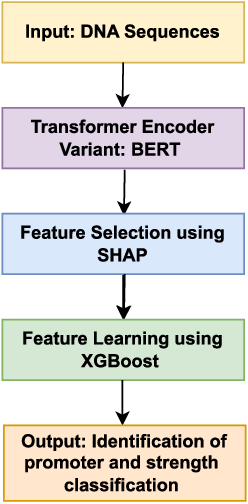}\hspace{2cm}
					\includegraphics[height=2.0in,width=1.2in]{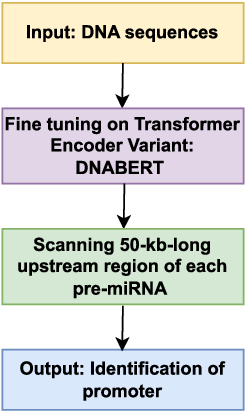}
				}\centerline{(a)\hspace{55mm}(b)}\vspace{1mm}
			        \centerline{
				\includegraphics[height=2.0in,width=4.0in]{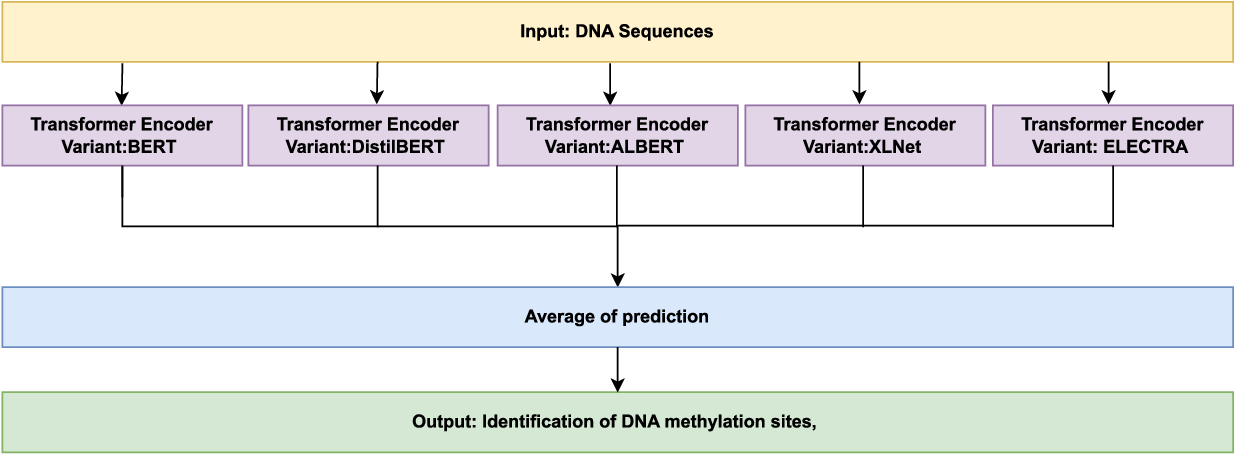}
				
					}
				\centerline{(c)}\vspace{1mm}
				\centerline{
                    \includegraphics[height=2.0in,width=1.2in]{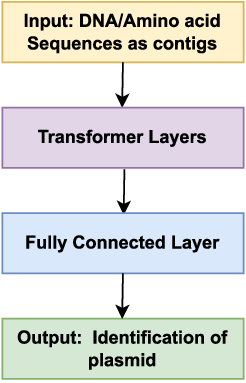}\hspace{0.2mm}
				\includegraphics[height=2.0in,width=1.2in]{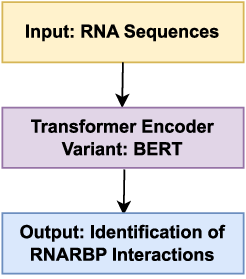}\hspace{0.2mm}
                    \includegraphics[height=2.0in,width=1.2in]{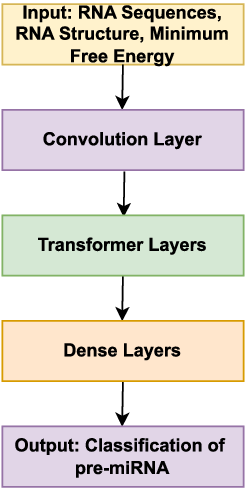}}
				\centerline{(d)\hspace{35mm}(e)\hspace{35mm}(f)}\vspace{1mm}
				\caption{Applications of Transformer-based language models to nucleotide sequences for (a) BERT-Promoter, (b) miProBERT, (c) MuLan-Methyl, (d) PLASMe, (e) BERT-RBP and (f) miRe2e}
				\label{nt}
			\end{figure}

A promoter region is a DNA sequence located upstream of the Transcription Starting Site of a gene containing specific motifs that proteins like Transcription Factors can bind to in order to regulate the gene transcription. It is a challenging task to identify promoter regions in order to understand the mechanisms that regulate gene expression.
Although several deep learning models based on CNN \cite{Le2019} and LSTM \cite{Oubounyt2019} have been proposed, the prediction performance is still not satisfactory. To address this problem, \cite{LE2022} has proposed a predictor based on BERT and SHapley Additive exPlanations (SHAP) \cite{lundberg2017} analysis along with machine learning algorithms. Initially, they have used BERT to extract features from DNA sequences and subsequently SHAP is used for reducing the dimension of features. Next, such features are used as inputs for different machine learning algorithms to predict DNA promoters as well as their strength. Among the several learning algorithms as used in this work, XGBoost provides the best results and is therefore selected as the desired machine learning algorithm (Fig. \ref{nt}(a)). In \cite{Wang2023}, Wang et al. have proposed miProBERT ((Fig. \ref{nt}(b))) which predicts miRNA promoters directly from gene sequences without considering any structural or biological signal. They have used the pretrained model DNABERT and subsequently finetuned it on the gene promoter dataset to scan the upstream regions of miRNA, thereby identifying 665 miRNA promoters. They have further used a random substitution strategy to create a negative dataset. miProBERT shows better performance than many other miRNA promoter prediction methods \cite{Liu2017, Rie2017} showing 78.13\% precision and 75.76\% recall. The authors have also verified the predicted promoter regions by analysing conservation, CpG content and histone marks. However, this work has certain weaknesses such as no biological experimental validation has been performed for the identified miRNA promoters. Also, due to the presence of alternative promoters it is possible for the same miRNA to have multiple promoters. Thus, the role of alternative promoters needs to be explored as well. Transformer models have also been used in \cite{Mai2022} for accurate promoter prediction of freshwater cyanobacterium Synechocystis sp. PCC 6803 and the fastest growing cyanobacterium Synechococcus elongatus sp. UTEX 2973. The promoters and non-promoters from Synechococcus elongatus sp. UTEX 2973 have been used to train the model which achieved an F1 score of 0.92 and an AUROC of 0.97. The validation has been performed with promoters from Synechocystis sp. PCC 6803, during which F1 score of 0.91 and AUROC score of 0.96 are achieved. The authors have further developed TSSNote-CyaPromBERT which is an integrated platform for dataset extraction, model training and also prediction of promoter from dRNA-seq datasets. Although the work shows quite competitive performance, it suffers from certain drawbacks. The model does not show very good results when tested on datasets of some other species. Also, the model works with sequential information thereby failing to capture complex interactions during transcription thereby leading to false positives in the regional scanning mode.Most of the works discussed so far have used a typical BERT model for the tasks. Such models rely on token prediction and may ignore domain knowledge such as repetitive elements, DNA binding sites etc. In \cite{Weizhi2022}, the authors have incorporated motif prediction along with token prediction to take care of such domain knowledge by using ELECTRA as the Transformer model. Promoter and transcription factor binding site predictions have also been explored in \cite{Weizhi2022} where the authors have proposed a self-supervised Motif-oriented DNA (MoDNA) pre-training framework that can be finetuned for many downstream tasks like promoter and TFBS prediction.  

On the other hand, enhancers are important \textit{cis}-regulatory elements regulating various biological functions as well as enhancing the transcription of target genes. In this regard, iEnhancer-ELM \cite{Li2023} which is based on BERT-like enhancer language models, have been used to tokenize DNA sequences with multi-scale \textit{k}-mers to extract contextual information through a multi-head attention mechanism. Initially, they have evaluated the performance of different scale \textit{k}-mers and later merged them for improved enhancer identification. The final classification is performed using a 2-layer perceptron network. The results show that iEnhancer-ELM scores better than methods which does not use BERT.But, this work does not differentiate between typical enhancers and super enhancers. Super enhancers are used for defining cell identity and regulating transcription of gene as well as they are associated with several known diseases. Usually, they span a large genomic domain by aggregating multiple typical enhancers together based on a specific stitching distance. Typical enhancers made up of a super enhancer are similar to normal enhancers on the genome and are involved in gene transcriptional regulation as well. Thus, it is important to distinguish super-enhancers from typical enhancers. In this regard, Luo et al. \cite{LUO2023} proposed SENet which is based on deep neural network model to discriminate between the two categories by using only sequence information. SENet uses dna2vec feature embedding, convolution for extracting local features, attention pooling for retaining finer features, and Transformer for extracting contextual information.
More recently Ligeti ~\cite{Ligeti2024} and colleagues proposed ProkBERT a family of language models for microbiome applications. It was successfully applied to promoter prediction and phage identification, demonstrating strong performances. ProkBERT models leverage transfer learning and self-supervised approaches to effectively handle complex microbial data. The innovative Local Context-Aware (LCA) tokenization enhances traditional transformers by preserving local context, enabling adaptability across diverse bioinformatics tasks.

\subsection{Methylation}
DNA methylation is a crucial biological process that primarily involves adding a methyl (CH3) group to the fifth carbon atom of a cytosine ring, but it can also occur through other mechanisms. It is often associated with many processes such as genomic imprinting, X-chromosome inactivation, repression of transposable elements, aging, and carcinogenesis.

Methylations drive epigenetic regulation of gene expression and have been also used as markers in metagenomic binning.
DNA methylation maps can be obtained either by using next-generation sequencing (NGS) approaches such as whole-genome bisulfite sequencing (WGBS) \cite{Lister2009} or reduced-representation bisulfite sequencing (RRBS) \cite{meissner2005}. However, these techniques are often costly and time-consuming \cite{flusberg2010}. Also, short-read sequencing prevents bisulfite sequencing from profiling DNA methylation in repetitive genomic regions \cite{landan2012, treangen2012}. These shortcomings have led to the development of computational based approaches such as Transformer-based language models to predict DNA methylation sites. In this regard, iDNA-ABF \cite{Jin2022} which is a multi-scale deep biological language learning model has been used to predict DNA methylations based on only genomic sequences. Most importantly, iDNA-ABF captures both sequential and functional semantics information from background genomes. For model construction, the authors have adopted DNABERT \cite{Ji2021dnabert}. Although, the work shows competent results, the authors neither considered multiple DNA modification sites for prediction nor used taxonomic information as explicit features. These shortcomings have been overcome in  \cite{Zeng2023} where the authors propose  model viz. MuLan-Methyl. It is based on five popular Transformer-based language models and aims to identify three types of DNA methylation sites; N6-adenine, N4-cytosine, and 5-hydroxymethylcytosine. The five models (BERT \cite{devlin2019}, DistilBERT \cite{sanh2020distilbert}, ALBERT \cite{lan2020albert}, XLNet \cite{Yang2019xlnet} and ELECTRA \cite{clark2020electra}) are combined together to collectively predict the DNA methylation status (Fig. \ref{nt}(c)). Furthermore, post-transcriptional 2'-O-methylation (Nm) RNA modification is crucial in various cellular tasks and related to a number of diseases such as hepatocellular carcinoma, lung adenocarcinoma and congenital muscular dystrophy. Techniques such as PCR-based approaches, RNaseH-based approaches \cite{Yu1997} and reverse transcription-based approaches \cite{Dong2012} are  few of the experimental approaches used to infer additional Nm modifications. However, they are slow thus paving the way for further computational methods such as Support Vector Machines \cite{CHEN2016, Yang2018}, Light Gradient Boosting Machines \cite{Ao2021}, BiLSTM with CNNs \cite{Li2021} etc. 
BERT2OME has been proposed by \cite{Soylu2023} et al. to predict Nm modification sites from RNA sequences where the model has been pretrained with RNA segments as input language. Once the pretraining is over, this high dimensional embedding dataset is fed into a 2D convolutional neural network to extract more attributes in order to finally predict RNA 2'-O-methylation modification sites over RNA sequences.
Stanojevi\'{c} \cite{Stanojevic2024} et al. have proposed Rockfish, a deep learning algorithm that significantly improves read-level 5-methylcytosine detection by using Nanopore sequencing. Rockfish demonstrates strong agreement with whole-genome bisulfite sequencing, requiring fewer reads while providing high confidence in CpG-rich regions. Its efficiency and performance in human and mouse samples make it a versatile tool for studying 5-methylcytosine methylation across species and diseases. Additionally, its adaptable design supports compatibility with evolving sequencing technologies and modification types.

\subsection{Reads}
In Next-Generation Sequencing (NGS) short sequenced fragments are called ``reads"; in most of the cases they are the product of DNA random shearing. Nowadays, the increase in NGS technologies provides a huge amount of data and the need for new methodologies to classify sample reads through machine learning approaches has become a priority. Finding an association between the microbiome and human health is an intriguing yet difficult topic that is feasible with the advent of next-generation sequencing (NGS) and metagenomic methods. Accurate identification is necessary to understand the connection between disease and the microbiome because even species that belong to the same genus might have diverse functions and pathogenicities.
In this regard, Gwak et al. \cite{Gwak2022} have applied Transformer-based embedding consisting of 12 embedding blocks to bacterial genomes in order to accurately identify species at the read-level. The model has been pretrained with Staphylococcus genomes and was finetuned for classifying species within the Staphylococcus genus. With the use of simulated reads of 151 and 251 bp, two distinct models have been trained as well wherein both models achieved an ROC-AUC values of over 0.98. Identification of Human Papillomavirus (HPV) reads in human host genomic data by using Transformer-based pipeline DeepViFi has been explored in \cite{Rajkumar2022}. DeepViFi has three components: (i) a Transformer for the production of latent representations of the short-reads, (ii) Classification of the latent representations using Random Forest classifier to determine if the read is HPV positive and (iii) a Light Gradient Boosting model for the sub-family identification of HPV. DeepViFi has achieved a precision-recall AUC of 0.94, 0.94, 0.91, and 0.16 for detecting HPV reads on the easy, intermediate, hard and non-human test sets respectively. It is important to note that though DeepViFi shows better performance than many existing methods, it does not consider domain-specific knowledge of viruses. In \cite{Gwak2022_1}, a hierarchical BERT model named ViBE has been used for the detection of eukaryotic viruses from metagenome sequencing data and classify them at the order level. ViBE has been adapted from BERT and DNABERT architecture and consists of an embedding and a classification layer. Both pre-training and finetuning are performed to classify the viruses. The pre-training step includes unsupervised learning of a large set of virus genomic sequences while the fine tuning involves the paired end reads generated from the viral and bacterial genome sequences. In their study, the authors have used three classifiers encompassing a domain-level classifier and two order-level classifiers for DNA and RNA viruses. Domain-level classifier is responsible for the screening of viral sequences while the order-level classifiers are two separate predictors for DNA and RNA viruses. With an AUROC of 0.98 ViBE performed much better when compared to other methods such as CHEER \cite{SHANG2021} and DVF \cite{ren2020}. 

Extracting information from short contigs is yet another area where Transformers can be effectively used. Tang et al. \cite{Tang2023} have developed PLASMe (Fig. \ref{nt} (d)) for plasmid detection using order-specific Transformer models. In this work, the authors have treated plasmids as a language defined on a vocabulary of proteins, thereby leveraging the use of Transformer to learn the importance of proteins and their associations for plasmid identification. To maximise the capacity of feature learning, 35 Transformer-based models are designed for plasmid detection. The authors have conducted the experiments both at the nucleotide and protein levels. Despite the desirable results shown by the authors in \cite{Tang2023}, the work has certain shortcomings. This work has considered only proteins of plasmids but not proteins specific to chromosomes. Considering such proteins may further improve the accuracy of the model. Also, predicting the functions of unannotated plasmid proteins may help the research community by studying the evolutionary as well as ecological significance of plasmids. A self-attention based deep learning metagenomic analysis tool viz. MetaTransformer \cite{Wichmann2023} has been proposed for the classification of metagenomic reads. The main motivation of the work is to improve the speed as well as the prediction accuracy of deep learning methodologies for classification. As a consequence, their model can be executed on single GPU as well. The authors have compared  MetaTransformer against DeepMicrobes \cite{Liangdeepmicrobes2020} as well as traditional \text{k}-mer based methods and the proposed model has been seen to outperform both. However, the model may be required to be trained in its entirety if a new species or genus is introduced.
 
\subsection{Binding}
Binding between molecules is a thermodynamically driven process, that in the case of nucleic acids can be investigated through primary sequences by applying a machine learning approach.
Cell functions which include mRNA modification, splicing, localisation and translation are regulated by interactions between RNA sequences and RNA-binding proteins (RBPs). In order to predict such interactions, statistical models especially support vector machine are widely used. Moreover, many deep learning models are also explored. However, the existing models lack interpretability and are mostly of complex nature leading to the development of new models. In this regard, Yamada et al. \cite{Yamada2022} have proposed BERT-RBP (Fig. \ref{nt} (e)) model which is pretrained on a human reference genome to predict the RBP-binding property of RNA sequences. Furthermore, attention analysis is applied on the finetuned model which shows that BERT-RBP is able to translate biological contexts from only RNA sequences (Fig. \ref{nt}(e)).

Identifying DNA-protein binding is another aspect where BERT can be specifically used. In \cite{HLuo2023}, Luo et al. have proposed TFBert which considers DNA sequences as natural sentences and \text{k}-mer nucleotides as words, thereby extracting upstream and downstream nucleotide information for 690 unlabeled ChIP-seq datasets. Experiments show that the average AUC is 94.7\% which outperforms most of the existing popular methods. Wang et al. \cite{Wang2023_1} have proposed SA-Net which uses \text{k}-mer (4-mer provided the best results) embedding to encode RNA sequences and thereafter utilises self-attention based neural network to extract sequence features. The experimental results show that SA-Net outperforms CNN and CNN-BLSTM models in sequence feature extraction. In the context of binding prediction, it is also a challenging task to predict which antigens a T-cell receptor may bind to. TCR-BERT has been proposed in \cite{Wu2021} which is a deep learning model with self-supervised transfer learning. TCR-BERT uses unlabeled TCR sequences to learn about TCR sequences and thereby may help in many downstream applications as well. Apart from being a useful tool for T-cell scientists, TCR-BERT also helps in solving challenging problems such as designing novel TCR sequences with engineered binding affinities. TF-DNA binding is explored in \cite{Zhang2022} where the authors have developed GHTNet (General Hybrid Transformer Network), a Transformer-based model to predict TF-DNA binding specificity. Transcription Factors (TFs) are proteins that bind to DNA and regulate gene expression. Prediction of TF-DNA binding is crucial for understanding how gene expression is controlled by TFs. GHTNet uses self-attention and CNN to predict TF-binding specificity. For encoding the DNA sequences, word2vec is employed while multi-head self-attention mechanism and convolution-FFN (C-FFN) are used to capture the global dependencies. Finally, multilayer perceptron (MLP) is utilized to detect high-level characteristics once CNN has finished extracting the low-level features of TFBSs. In \cite{Ding2023}, Ding et al. have used a deep learning architecture viz. DeepSTF for the prediction of TFBS by combining DNA sequence and shape profiles. In this regard, the higher order DNA sequence features are extracted by stacked CNN while the DNA shape profiles are extracted by integrating improved Transformer encoder and Bi-LSTM. Subsequently, the higher order sequence features and the shape profiles are considered together to identify TFBSs. In their work, the authors have explained the use of DNA sequence and shape to understand and learn the important features. Moreover, they have incorporated multi-head attention mechanism to learn high level representation of input data in a different representation space. Their results demonstrate a superior performance as compared to existing methods for the prediction of TFBSs. TFBS prediction has also been explored in~\cite{Ghosh2024} where the authors have used DNABERT for embedding and eventually capsule network along with bidirectional long short term memory for predicting TFBS. Along with traditional prediction, the authors have also performed cross-cell line prediction to show the generalisabilty of the proposed model. The results show that the proposed method performs better than existing state -of-the-art models such as DeepARC, DeepTF, CNN-Zeng and DeepBind.

Although, the aforementioned works have achieved competent performances, there are scopes of improvement. Most of the works have used DNA sequences and shapes for prediction of TF-DNA bindings. However, various other factors such as chromatin structure, nucleosome occupancy and chemical modifications are also important to understand and predict such binding.


\subsection{Miscellaneous}
A generalised pretrained large language model DNAGPT which is trained on over 200 billion base pairs from all mammals is proposed in \cite{zhang2023dnagpt}. The authors have improved upon the classical GPT (Generative PreTrained Transformer) model by incorporating binary classification task of DNA sequence order and a numerical regression task of guanine-cytosine content prediction. The model behaves as a comprehensive token language as well. Thus, DNAGPT can work with different DNA analysis tasks while processing both sequence and numerical data. By comparing with many state-of-the-art models, the authors have proven the superiority of DNAGPT in genomic signal, region recognition, mRNA abundance regression and artificial genome generation task. 
Fishnman et al. \cite{Fishman2023} have proposed GENA-LM which is a suite of Transformer-based foundational DNA language models that can handle input lengths of up to 36000 base pairs. GENA-LM is made up of bert-base, bert-base-t2t, bert-base-lastln-t2t, bert-base-t2t-multi and bert-large-t2t. For the sequence tokenisation, the authors have used byte-pair encoding (BPE) where the dictionary size is set to 32000. Once the model is ready, the authors have used it for several downstream tasks such as prediction of promoters, splice site, drosophila enhancers and polyadenylation sites as well as chromatin profiling. However, the authors have tested their model against those sequences which can exploit the benefits of a longer context. Such tests show that GENA-LM outperforms other pretrained models including task-specific CNNs. Nucleotide Transformer is proposed in \cite{Dalla-Torre2023} which is pretrained on DNA sequences while integrating information from 3202 diverse human genomes as well as 850 genomes from a diverse range of species. The experimental results show that the model has commendable performance in 15 out of 18 downstream tasks using finetuning. DNABERT-2 which is an improvement over DNABERT is proposed in \cite{zhou2023}. DNABERT-2 uses byte-pair encoding and is pretrained with masked language modelling loss with a mask ratio of 15\%. It has then been used for core promoter detection as well as other downstream tasks. The authors have used Genome Understanding Evaluation (GUE) benchmark, which includes 7 genome sequence classification problems with 28 datasets. The results indicate that DNABERT-2 outperforms both DNABERT and nucleotide Transformers. TIS Transfomer has been proposed by Clauwaert et al. \cite{Clauwaert2023} to determine translation start sites by considering the information in the transcript nucleotide sequence. They showed that the limitations in the performance of TIS Transformer  are due to the absence of high-quality annotations and not due to the model itself. However, the advantages of the model are the ability to detect key features of the translation process and multiple coding sequences on a transcript. 
For the prediction of cross-immunity between viral strains, Du et al. \cite{du2023} have proposed DNA Pretrained Cross-Immunity
Protection Inference (DPCIPI) model. Such prediction is important for public health surveillance as well as vaccine development. In this regard, the authors have used DNABERT to create the initial encoding which is then followed by BiLSTM to capture the sequence meaning. Finally, a multi-layer perceptron neural network (MLP) is considered to get the final classification result. The authors have compared their model with several other statistical methods and DPCIPI outperfoms all of them for both binary and multi-level cross immunity prediction. Bai et al. \cite{Bai2022} have proposed IdentificatioN of bacteriopHagEs using deep RepresentatIon model with pre-Training (INHERIT) to identify bacteriophages which are viruses that infect and replicate within bacteria and archaea and rich in human body. Researchers can more effectively investigate phages by employing a technique that can accurately discriminate between phages and bacteria. INHERIT is an integrated model based on DNABERT and with an F1 score of 0.9932 has shown best performance when compared to state-of-the-art methods like VIBRANT \cite{Kieft2020}, VirSorter2 \cite{Guo2021_sorter}, Seeker \cite{Auslander2020} and DeepVirFinder \cite{ren2020}. A Transformer-based, three-dimensional chromatin conformation-aware deep learning architecture Chromoformer is proposed in \cite{Lee2022} for the quantitative decipher of histone codes in gene regulation. Gene expression is controlled by diverse group of regulators encompassing transcription factors, coactivators and corepressors and histone modifications (HMs) play the key role in the interplay among these factors. Chromoformer is trained on seven major HMs including H3K4me1, H3K4me3, H3K9me3, H3K27me3, H3K36me3, H3K27ac, and H3K9ac. It consists of three independent modules each accepting input features at different resolutions and producing an embedding vector. These three embeddings are then concatenated and fed into fully-connected layers to predict gene expression. Prediction of miRNAs is yet another area where the application of Transformers cannot be overlooked. miRe2e (Fig. \ref{nt}(f)) as proposed in \cite{Raad2021} is the first full end-to-end deep learning model for pre-miRNA prediction and is based on Transformers (Fig. \ref{nt}(f)). The model can accept raw genome-wide data as input, without any pre-processing or feature engineering. miRe2e has been tested with many experimental setups with human genome and has shown much improved performance as compared to state-of-the-art methods. Prediction of RNA modifications is carried out in \cite{Zhang2023} by employing MRM-BERT (Multiple kinds of RNA Modifications by BERT) which works by feeding task-specific sequences into BERT. MRM-BERT can predict multiple RNA modifications such as  pseudouridine, m6A, m5C, and
m1A in Mus musculus, Arabidopsis thaliana, and Saccharomyces
cerevisiae. The advantage of MRM-BERT lies in the fact that it follows the idea of `one-model-does-it-all' where it becomes easier to extend the model to predict various RNA modifications. The model has shown competitive performance on eight datasets. Transformers have also been applied to cancer data for tumour type classification based on per-patient omics data-sets. In this regard, Jurenaite et al. \cite{Jurenaite2022} have proposed SETQUENCE which is a Transformer-based deep neural network for supervised learning of oncology related tasks. The model has also been extended for multiple sources of omics data with SETOMIC. SETQUENCE and SETOMIC use DNABERT to encode the mutations while a fully connected layer with a rectified linear unit is used for classification. Avsec et al. \cite{Avsec2021}
have used deep learning architecture based on Transformer viz. Enformer for predicting gene expression and chromatin states from DNA sequences.  Moreover, Enformer has also been trained to predict enhancer-promoter interactions directly from DNA sequences. Training and testing of the model are performed on human and mouse genomes where improved prediction is observed when compared to existing similar predictor like Basenji2 \cite{Kelley2020}. Recently, taking inspiration from Hyena, which is based on implicit convolutions, Nguyen has proposed HyenaDNA \cite{Nguyen2023} which is a genomic foundation model pretrained on the human reference genome with context lengths of up to 1 million tokens at the single nucleotide level. Owing to the sub-quadratic scaling of sequence length, HyenaDNA trains 160 times faster than Transformer. It uses single nucleotide tokens and has complete global context at each layer. When compared with Nucleotide Transformer, HyenaDNA performs better on 12 of 18 datasets of downstream tasks by using a model with 1500 times fewer parameters and 3200 times less pretraining data. However, HyenaDNA uses only one human reference genome for pretraining which can lead to increased bias and generalisation capacity on learned features can also get diminished. 

All the aforementioned works focus mostly on DNA sequences while considering multi-omics and spatial-omics data may greatly increase the applicability of the proposed methods.

\section{Challenges and Future Directions}
Transformer-based language models have been proven to provide effective and significant results when applied to biological sequences, especially for their ability to define and handle a huge number of context-dependent features. 
Nevertheless, in order to build even more reliable and better performing Transformer models, some issues are still to be addressed; a huge amount of computational resources is needed to build Transformer models. Thus, the scientific community is devoting many efforts in order to reduce the computational load both in terms of time and space complexity. Also, some features of the models can be customized and particular attention could be paid to overcome the common limitation of deep learning model in interpreting and reading the intrinsic meaning of the models. There is still a large margin of improvement that could be achieved by developing models tailored for specific contexts as well as focus can be given on decoder-based models .
Moreover, features apart from DNA sequences can be explored in order to come up with better performance results.

\begin{itemize}
    \item Computationally expensive: Although the main advantage of Transformers come from the self-attention module, it also leads to a very high computational expense which increases quadratically with the input sequence length, thereby making Transformer not able to model long sequences. Some of the works which have tried to mitigate or improve the Transformer for this problem are discussed below:
    \begin{itemize}
        \item Improvement in self-attention module: Beltagy et al. \cite{beltagy2020} have introduced local windowed attention with task motivated global attention in place of standard self-attention. This enables processing of longer sequences. BIGBIRD as proposed in \cite{zaheer2021big} have used a sparse attention mechanism to reduce the quadratic dependency on sequence length to linear. Performers introduced in \cite{choromanski2022} have used linear attention by replacing softmax with another approach, thereby using only linear space and time complexity. However, all these works have focussed on customising the self-attention module, thus retraining the whole Transformer model. 
        
        \item Dividing long sequences into chunks: Xie et al. \cite{xie2023chunk} have proposed dividing the long input sequences into a batch of chunks of feasible lengths and then selecting the most relevant tokens for decoding. 

         \item Few-shot learning: Few-shot learning is used for learning new tasks when provided with only a handful of data. In this regard, Transformers models may be adapted to new tasks using only a small set of parameters for nucleotide sequences, thereby being computationally efficient.

    \end{itemize}
    \item Model interpretability: Deep learning models are known for their lack of interpretability. However, in the field of computational biology and bioinformatics, interpretability is of utmost important in order to gain insights from the model. In this regard, the self-attention mechanism of Transformers is very important. For example, the attention mechanisms in DNABERT \cite{Ji2021dnabert} and DNABERT-2 \cite{zhou2023} can focus on important areas for decision-making, thus providing an interpretability to the models by conveying semantic meaning. Both the aforementioned models can rank the input nucleotide molecules as well as find the relationship between the input sequence contexts. Clauwaert et al. \cite{Clauwaert2023} have built a model from Transformer-XL to detect and characterise transcription factor starting sites, thereby showing the potential of Transformers in extracting biological meaning.

    \item Creating specific pretrained models for different contexts (at genus, family or species level or for metagenomics analysis): One very important direction for future work can be developing specific models that can handle data at genus, family, species level for metagenomic analysis. Metagenomic reads have shown to perform well for eukaryotic viral taxa at the order level \cite{Gwak2022}. Such work can also be extended for prediction at genus level. Moreover, most of the works have used \text{k}-mer representations for the sequence data. However, as pointed out in \cite{zhou2023}, overlapping \text{k}-mer tokenisation may result in information leakage in masked language modelling. Non-overlapping \text{k}-mer tokenisation on the other hand may not be very efficient. To mitigate the aforementioned problems, Byte-Pair Encoding \cite{sennrich2016} can be considered to be a suitable replacement for tokenisation in Transformers. 

    \item Generation-based models: Most of the works discussed in this review are encoder-based models, while few focus on decoder-based models as well. However, generation-based models is by large under investigated in bioinformatics.

    \item Multi-modal pre-training: Instead of only considering DNA sequences, making use of other factors such as shape, chromatin structure, nucleosome occupancy and chemical modification are beneficial while pre-training models, thereby leading to improved performance.

\end{itemize}
\section{Conclusion}
The revolution of Transformer-based models is unprecedented in the field of natural language processing. Such models have also brought new waves in bioinformatics as well, particularly for the processing of nucleotide sequences. In this regard, BERT-promoter with an accuracy of 85.5\% for promoter identification, miProBERT showing recall of 75.76\% for predicting miRNA promoters etc. are some of the examples which show how Transformer-based models can be utilised for several purposes in bioinformatics. Such effective and reliable performances pave the way for the application of Transformer-based models to many different fields encompassing nucleotide sequences. For instance, they can be applied for the recognition of stable nucleosome forming nucleotide sequences, as well as in immunoinformatics applied for the identification of potential epitopes.
Transformer-based models seem to be particularly suitable for designing models that are able to identify Transcription Factor Binding Sites for an individual TF.  However, application of Transformers in bioinformatics and related fields are still in the nascent stage. 
Better pretrained models, combining Transformer models with other newly developed deep learning techniques as well as improving model flexibility are some areas which can be focused on. We hope that this review article helps the researchers working in the field of bioinformatics to propose new and improved methodologies to mitigate various problems pertaining to the diagnosis and treatment of human diseases. 

\begin{longtable}{p{2cm}p{2cm}p{8cm}p{2cm}}
       Category  & Paper & Main Idea &  Data Repository \\\hline
       Promoter and Enhancer & Le et al. \cite{LE2022} & Pre-trained BERT model is used to encode DNA  sequences while feature selection using SHAP analysis is performed to select the top-rank BERT encodings to provide them as input to machine learning algorithms for promoter prediction
       & 
       \href{https://github.com/khanhlee/bert-promoter}{BERT-Promoter}\\\cline{2-4}
       &Wang et al. \cite{Wang2023} & A BERT-based model for predicting miRNA promoters directly from gene sequences without using any structural or biological signals   &
       \href{https://github.com/xwang1427/miProBERT}{miProBERT}\\\cline{2-4} 
       &Mai et al. \cite{Mai2022} &Comparison of performance of state-of-the-art NLP models to predict and analyze promoters in cyanobacterium Synechocystis sp. PCC 6803 and cyanobacterium Synechococcus elongatus sp. UTEX 2973&
       \href{https://github.com/chenli-bioinfo/promoter}{TSSNote}\\\cline{2-4} 

       & An et al. \cite{Weizhi2022} & Motif-Oriented DNA pre-training framework based on self-supervised design as well as can be finetuned for predicting promoters and TFBS& N/A\\\cline{2-4}

       & Li et al. \cite{Li2023} & iEnhancer-ELM tokenizes DNA sequences with multi-scale \text{k}-mers and extracts contextual information of different scale \text{k}-mers  related with their positions via multi-head attention mechanism for enhancer prediction
   &
       \href{https://github.com/chen-bioinfo/iEnhancer-ELM}{iEnhancer-ELM}\\\cline{2-4} 
       
       & Luo et al. \cite{LUO2023}& A deep learning framework for discriminating  super- and typical enhancers by sequence information & 
       \href{https://github.com/lhy0322/SENet}{SENet}\\\hline

       Methylation 
        
       & Zeng et al. \cite{Zeng2023} &  Prediction of DNA methylation sites using 5 Transformer based languages& \href{https://github.com/husonlab/mulan-methyl}{MuLan-Methyl}\\\cline{2-4}
       & Jin et al. \cite{Jin2022} & Enables the interpretable prediction of DNA methylations based on genomic sequences only  & \href{https://github.com/FakeEnd/iDNA_ABF}{iDNA-ABF}\\\cline{2-4}
       &Soylu et al. \cite{Soylu2023}& Understanding post-transcriptional 2'-O-methylation (Nm) RNA modification using BERT-based model and CNN&
       \href{https://github.com/seferlab/bert2ome}{BERT2OME}\\\hline
       Reads & Gwak et al.  \cite{Gwak2022} & Classify species based on whole-genome sequencing reads & N/A\\\cline{2-4} 
       & Rajkumar et al. \cite{Rajkumar2022} & Transformer based pipeline to detect viral reads in short-read whole genome sequence data & N/A \\\cline{2-4} 
       &Gwak et al. \cite{Gwak2022_1} &A hierarchical BERT model to identify eukaryotic viruses using metagenome sequencing data
       & \href{https://github.com/DMnBI/ViBE}{ViBE}\\\cline{2-4} 
       & Tang et al. \cite{Tang2023} & Identifying PLASMid contigs from short-read assemblies using Transformer& \href{https://github.com/HubertTang/PLASMe}{PLASMe}\\\cline{2-4} 
       & Wichmann et al. \cite{Wichmann2023} & Classifying reads using Transformers & N/A\\\hline 
       
       Binding 
   
       & Yamada et al. \cite{Yamada2022} &Prediction of RNA–protein interactions by using BERT model pretrained on a human reference genome& \href{https://github.com/kkyamada/bert-rbp}{BERT-RBP}\\\cline{2-4} 
        & Luo et al. \cite{HLuo2023} & DNA–protein binding prediction based on task-specific pre-training& \href{https://github.com/lhy0322/TFBert}{TFBert}\\\cline{2-4} 
       &Wang et al. \cite{Wang2023_1} &Self-Attention based neural network to predict RNA-Protein binding sites & \href{https://github.com/aliezxy/SA-Net}{SA-Net}\\\cline{2-4} 
       & Wu et al. \cite{Wu2021} & Using BERT to predict TCR-antigen binding & \href{https://github.com/wukevin/tcr-bert}{TCR-BERT}\\\cline{2-4} 
      & Zhang et al. \cite{Zhang2022} & Using a hybrid Transformer network to predict TF-DNA binding specificity &\href{https://github.com/ZhangLab312/GHTNet}{GHTNet}\\\hline
    Miscellaneous 
    & Zhang et al. \cite{zhang2023dnagpt} & A Generalized pretrained tool for multiple DNA sequence analysis tasks & N/A \\\cline{2-4} 
     & Fishman et al. \cite{Fishman2023}& Open-Source foundational models for applications in long DNA sequences & \href{https://github.com/AIRI-Institute/GENA_LM}{GENA-LM}\\\cline{2-4} 
      
     & Dalla-Torre et al. \cite{Dalla-Torre2023} & Building and applying nucleotide Transformer for several downstram tasks & N/A \\\cline{2-4} 
             & Zhou et al. \cite{zhou2023} & Proposal of a foundation model pre-trained on multi-species genomes& \href{https://github.com/Zhihan1996/DNABERT_2}{DNABERT-2}\\\cline{2-4}
    & Clauwaert et al. \cite{Clauwaert2023} & A deep learning model for the determination of translation start sites solely utilizing the information embedded in the transcript nucleotide sequence&
       \href{https://github.com/jdcla/TIS_Transformer}{TIS Transformer}\\\cline{2-4}
     & Du et al. \cite{du2023} & A pre-trained deep learning model for estimation of cross-immunity between drifted strains of Influenza A/H3N2 & \href{https://github.com/Elvin-Yiming-Du/DPCIPI_cross-immunity_prediction}{DPCIPI}\\\cline{2-4} 
    & Bai et al. \cite{Bai2022} & Identification of bacteriophage genome sequences with representation learning & \href{https://github.com/Celestial-Bai/INHERIT}{INHERIT}\\\cline{2-4} 
    & Lee et al. \cite{Lee2022} & Learning the histone codes with large genomic windows and three-dimensional chromatin interactions using Transformer & \href{https://github.com/dohlee/chromoformer}{chromoformer}\\\cline{2-4} 
    & Raad et al. \cite{Raad2021} & A full end-to-end deep model based on Transformers for prediction of pre-miRNAs &
    \href{https://github.com/sinc-lab/miRe2e}{miRe2e}\\\cline{2-4} 
    & Zhang et al. \cite{Zhang2023} & Prediction of multiple types of RNA modifications via biological language model  & \href{http://csbio.njust.edu.cn/bioinf/mrmbert/}{Mrmbert}\\\cline{2-4} 
    & Jurenaite et al. \cite{Jurenaite2022} & Supervised learning of oncology related tasks & N/A \\\cline{2-4} 
    & Avsec et al. \cite{Avsec2021} & Predicting gene expression and chromatin states in humans and mice from DNA sequences & \href{https://github.com/deepmind/deepmind-research/tree/master/enformer}{Enformer}\\\hline
    
\caption{Brief description of representative applications of Transformer-based language models for Nucleotide sequences}

\label{tab:1}
    
\end{longtable}

\section*{Ethics approval and consent to participate}
The ethical approval or individual consent was not applicable.

\section*{Consent for publication}
Not applicable.

\section*{Funding}
This work has been carried out under SERB-SURE scheme, funded by Government of India. This work is also supported by PNRR MUR project PE0000013-\textbf{FAIR}, Next Generation EU program, Italian Government.

\section*{Conflict of interest}
The authors declare that they have no conflict of interest.

\section*{Author contributions}
\textbf{Nimisha Ghosh}: Conceptualization; Visualization; Writing - original draft \& editing, \textbf{Daniele Santoni}: Visualization; Writing - original draft \& editing, \textbf{Indrajit Saha}: Writing - original draft \& editing, 
\textbf{Giovanni Felici}: Supervision; Funding acquisition; Project administration; Resources; Validation; Visualization; Writing - review \& editing.

\section*{Acknowledgment}
Daniele Santoni is a member of the National Group for Scientific Computing (GNCS) - Istituto di Alta Matematica Francesco Severi (INdAM). The authors would also like to thank Dr. Ajit Kumar Nayak for his valuable suggestions.

\bibliographystyle{elsarticle-num}
\bibliography{References}

\begin{thebibliography}{100}
\expandafter\ifx\csname url\endcsname\relax
  \def\url#1{\texttt{#1}}\fi
\expandafter\ifx\csname urlprefix\endcsname\relax\def\urlprefix{URL }\fi
\expandafter\ifx\csname href\endcsname\relax
  \def\href#1#2{#2} \def\path#1{#1}\fi

\bibitem{BEACH1943}
E.~F. Beach, B.~Munks, A.~Robinson, The amino acid composition of animal tissue
  protein, Journal of Biological Chemistry 148~(2) (1943) 431--439.
\newblock \href {https://doi.org/10.1016/S0021-9258(18)72300-4}
  {\path{doi:10.1016/S0021-9258(18)72300-4}}.

\bibitem{Sanger49}
F.~Sanger, {The terminal peptides of insulin}, Biochemical Journal 45~(5)
  (1949) 563--574.
\newblock \href {https://doi.org/10.1042/bj0450563}
  {\path{doi:10.1042/bj0450563}}.

\bibitem{Ferran1991}
E.~A. Ferr{á}n, P.~Ferrara, Topological maps of protein sequences, Biological
  Cybernetics 65 (1991) 451--458.
\newblock \href {https://doi.org/doi.org/10.1007/BF00204658}
  {\path{doi:doi.org/10.1007/BF00204658}}.

\bibitem{Orengo1993}
C.~Orengo, T.~Flores, W.~Taylor, et~al., {Identification and classification of
  protein fold families}, Protein Engineering, Design and Selection 6~(5)
  (1993) 485--500.
\newblock \href {https://doi.org/10.1093/protein/6.5.485}
  {\path{doi:10.1093/protein/6.5.485}}.

\bibitem{blekas2005}
K.~Blekas, D.~I. Fotiadis, A.~Likas, Motif-based protein sequence
  classification using neural networks, Journal of Computational Biology 12~(1)
  (2005) 64--82.
\newblock \href {https://doi.org/10.1089/cmb.2005.12.64}
  {\path{doi:10.1089/cmb.2005.12.64}}.

\bibitem{Exarchos2006}
T.~P. Exarchos, C.~Papaloukas, C.~Lampros, et~al., Protein classification using
  sequential pattern mining, in: 2006 International Conference of the IEEE
  Engineering in Medicine and Biology Society, 2006, pp. 5814--5817.
\newblock \href {https://doi.org/10.1109/IEMBS.2006.260336}
  {\path{doi:10.1109/IEMBS.2006.260336}}.

\bibitem{Kocsor2005}
A.~Kocsor, A.~Kert{é}sz-Farkas, L.~Kaj{á}n, et~al., {Application of
  compression-based distance measures to protein sequence classification: a
  methodological study}, Bioinformatics 22~(4) (2005) 407--412.
\newblock \href {https://doi.org/10.1093/bioinformatics/bti806}
  {\path{doi:10.1093/bioinformatics/bti806}}.

\bibitem{karlin1997}
S.~Karlin, J.~Mrazek, A.~M. Campbell, Compositional biases of bacterial genomes
  and evolutionary implications, Journal of bacteriology 179~(12) (1997)
  3899--3913.
\newblock \href {https://doi.org/10.1128/jb.179.12.3899-3913.1997}
  {\path{doi:10.1128/jb.179.12.3899-3913.1997}}.

\bibitem{Karlin97}
S.~Karlin, J.~Mrazek, Compositional differences within and between eukaryotic
  genomes, in: Proceedings of the National Academy of Sciences, Vol.~94, 1997,
  p. 10227–10232.
\newblock \href {https://doi.org/10.1073/pnas.94.19.10227}
  {\path{doi:10.1073/pnas.94.19.10227}}.

\bibitem{IUCHI2021}
H.~Iuchi, T.~Matsutani, K.~Yamada, et~al., Representation learning applications
  in biological sequence analysis, Computational and Structural Biotechnology
  Journal 19 (2021) 3198--3208.
\newblock \href {https://doi.org/10.1016/j.csbj.2021.05.039}
  {\path{doi:10.1016/j.csbj.2021.05.039}}.

\bibitem{Zeng2018}
W.~Zeng, M.~Wu, R.~Jiang, {Prediction of enhancer-promoter interactions via
  natural language processing}, BMC Genomics 19~(84) (2018).

\bibitem{Blacoe2012}
W.~Blacoe, M.~Lapata, A comparison of vector-based representations for semantic
  composition, in: Proceedings of the 2012 Joint Conference on Empirical
  Methods in Natural Language Processing and Computational Natural Language
  Learning, 2012, p. 546–556.

\bibitem{Turian2010}
J.~Turian, L.-A. Ratinov, Y.~Bengio, Word representations: A simple and general
  method for semi-supervised learning, in: Proceedings of the 48th Annual
  Meeting of the Association for Computational Linguistics, 2010, pp. 384--394.

\bibitem{Mikolov2013a}
T.~Mikolov, I.~Sutskever, K.~Chen, et~al., Distributed representations of words
  and phrases and their compositionality, in: Advances in Neural Information
  Processing Systems, Vol.~26, 2013.

\bibitem{Mikolov2013b}
T.~Mikolov, K.~Chen, G.~Corrado, et~al., Efficient estimation of word
  representations in vector space (2013).
\newblock \href {http://arxiv.org/abs/1301.3781} {\path{arXiv:1301.3781}}.

\bibitem{Schuster1997}
M.~Schuster, K.~Paliwal, Bidirectional recurrent neural networks, IEEE
  Transactions on Signal Processing 45~(11) (1997) 2673--2681.
\newblock \href {https://doi.org/10.1109/78.650093}
  {\path{doi:10.1109/78.650093}}.

\bibitem{Hochreiter1997}
S.~Hochreiter, J.~Schmidhuber, Long short-term memory, Neural Comput. 9~(8)
  (1997) 1735–1780.
\newblock \href {https://doi.org/10.1162/neco.1997.9.8.1735}
  {\path{doi:10.1162/neco.1997.9.8.1735}}.

\bibitem{cho2014}
K.~Cho, B.~van Merri{\"e}nboer, C.~Gulcehre, et~al., Learning phrase
  representations using {RNN} encoder{--}decoder for statistical machine
  translation, in: Proceedings of the 2014 Conference on Empirical Methods in
  Natural Language Processing ({EMNLP}), 2014, pp. 1724--1734.
\newblock \href {https://doi.org/10.3115/v1/D14-1179}
  {\path{doi:10.3115/v1/D14-1179}}.

\bibitem{Zeng2020review}
Y.~Zeng, M.~Gong, M.~Lin, et~al., A review about transcription factor binding
  sites prediction based on deep learning, IEEE Access 8 (2020) 219256--219274.
\newblock \href {https://doi.org/10.1109/ACCESS.2020.3042903}
  {\path{doi:10.1109/ACCESS.2020.3042903}}.

\bibitem{NIPS2017}
A.~Vaswani, N.~Shazeer, N.~Parmar, et~al., Attention is all you need, in:
  Advances in Neural Information Processing Systems, Vol.~30, 2017.

\bibitem{Dai2019}
Z.~Dai, Z.~Yang, Y.~Yang, et~al., Transformer-xl: Attentive language models
  beyond a fixed-length context, in: Proceedings of the 57th Conference of the
  Association for Computational Linguistics, {ACL} 2019, Florence, Italy, July
  28- August 2, 2019, Volume 1: Long Papers, 2019, pp. 2978--2988.
\newblock \href {https://doi.org/10.18653/v1/p19-1285}
  {\path{doi:10.18653/v1/p19-1285}}.

\bibitem{zhao2019}
G.~Zhao, J.~Lin, Z.~Zhang, et~al., Explicit sparse transformer: Concentrated
  attention through explicit selection (2019).
\newblock \href {http://arxiv.org/abs/1912.11637} {\path{arXiv:1912.11637}}.

\bibitem{Kitaev2020}
N.~Kitaev, L.~Kaiser, A.~Levskaya, Reformer: The efficient transformer, in:
  International Conference on Learning Representations, 2020.

\bibitem{beltagy2020}
I.~Beltagy, M.~E. Peters, A.~Cohan, Longformer: The long-document transformer
  (2020).
\newblock \href {http://arxiv.org/abs/2004.05150} {\path{arXiv:2004.05150}}.

\bibitem{HAN2021ai}
X.~Han, Z.~Zhang, N.~Ding, et~al., Pre-trained models: Past, present and
  future, AI Open 2 (2021) 225--250.
\newblock \href {https://doi.org/10.1016/j.aiopen.2021.08.002}
  {\path{doi:10.1016/j.aiopen.2021.08.002}}.

\bibitem{radford2018}
A.~Radford, K.~Narasimhan, T.~Salimans, et~al., Improving language
  understanding by generative pre-training, OpenAI (2018).

\bibitem{radford2019}
A.~Radford, J.~Wu, R.~Child, et~al., Language models are unsupervised multitask
  learners, OpenAI blog 1~(8) (2019) 9.

\bibitem{Brown2020}
T.~Brown, B.~Mann, N.~Ryder, et~al., Language models are few-shot learners, in:
  Advances in Neural Information Processing Systems, Vol.~33, 2020, pp.
  1877--1901.

\bibitem{devlin2019}
J.~Devlin, M.-W. Chang, K.~Lee, et~al., {BERT}: Pre-training of deep
  bidirectional transformers for language understanding, in: Proceedings of the
  2019 Conference of the North {A}merican Chapter of the Association for
  Computational Linguistics: Human Language Technologies, Volume 1 (Long and
  Short Papers), 2019, pp. 4171--4186.
\newblock \href {https://doi.org/10.18653/v1/N19-1423}
  {\path{doi:10.18653/v1/N19-1423}}.

\bibitem{liu2019roberta}
Y.~Liu, M.~Ott, N.~Goyal, et~al., {RoBERTa}: A robustly optimized bert
  pretraining approach (2019).
\newblock \href {http://arxiv.org/abs/1907.11692} {\path{arXiv:1907.11692}}.

\bibitem{Yang2019xlnet}
Z.~Yang, Z.~Dai, Y.~Yang, et~al., XLNet: Generalized Autoregressive Pretraining
  for Language Understanding, 2019.

\bibitem{zhang2019ernie}
Z.~Zhang, X.~Han, Z.~Liu, et~al., {ERNIE}: Enhanced language representation
  with informative entities, in: Proceedings of the 57th Annual Meeting of the
  Association for Computational Linguistics, 2019, pp. 1441--1451.
\newblock \href {https://doi.org/10.18653/v1/P19-1139}
  {\path{doi:10.18653/v1/P19-1139}}.

\bibitem{Raffel2020t5}
C.~Raffel, N.~Shazeer, A.~Roberts, et~al., Exploring the limits of transfer
  learning with a unified text-to-text transformer, The Journal of Machine
  Learning Research 21~(140) (2020) 5485–5551.

\bibitem{lan2020albert}
Z.~Lan, M.~Chen, S.~Goodman, et~al., {ALBERT}: A lite {BERT} for
  self-supervised learning of language representations (2020).
\newblock \href {http://arxiv.org/abs/1909.11942} {\path{arXiv:1909.11942}}.

\bibitem{Zhang2023_1}
S.~Zhang, R.~Fan, Y.~Liu, et~al., {Applications of transformer-based language
  models in bioinformatics: a survey}, Bioinformatics Advances 3~(1) (2023)
  vbad001.
\newblock \href {https://doi.org/10.1093/bioadv/vbad001}
  {\path{doi:10.1093/bioadv/vbad001}}.

\bibitem{Martinek2022}
V.~Martinek, D.~Cechak, K.~Gresova, et~al., Fine-tuning transformers for
  genomic tasks, bioRxiv (2022).
\newblock \href {https://doi.org/10.1101/2022.02.07.479412}
  {\path{doi:10.1101/2022.02.07.479412}}.

\bibitem{Le2019}
N.~Q.~K. Le, E.~K.~Y. Yapp, N.~Nagasundaram, et~al., Classifying promoters by
  interpreting the hidden information of dna sequences via deep learning and
  combination of continuous fasttext n-grams, Frontiers in Bioengineering and
  Biotechnology 7 (2019).
\newblock \href {https://doi.org/10.3389/fbioe.2019.00305}
  {\path{doi:10.3389/fbioe.2019.00305}}.

\bibitem{Oubounyt2019}
M.~Oubounyt, Z.~Louadi, H.~Tayara, et~al., Deepromoter: Robust promoter
  predictor using deep learning, Frontiers in Genetics 10 (2019).
\newblock \href {https://doi.org/10.3389/fgene.2019.00286}
  {\path{doi:10.3389/fgene.2019.00286}}.

\bibitem{LE2022}
N.~Q.~K. Le, Q.-T. Ho, V.-N. Nguyen, et~al., Bert-promoter: An improved
  sequence-based predictor of dna promoter using bert pre-trained model and
  shap feature selection, Computational Biology and Chemistry 99 (2022) 107732.
\newblock \href {https://doi.org/10.1016/j.compbiolchem.2022.107732}
  {\path{doi:10.1016/j.compbiolchem.2022.107732}}.

\bibitem{lundberg2017}
S.~M. Lundberg, S.-I. Lee, A unified approach to interpreting model
  predictions, Advances in neural information processing systems 30 (2017).

\bibitem{Wang2023}
X.~Wang, X.~Gao, G.~Wang, et~al., {miProBERT: identification of microRNA
  promoters based on the pre-trained model BERT}, Briefings in Bioinformatics
  24~(3) (2023) bbad093.
\newblock \href {https://doi.org/10.1093/bib/bbad093}
  {\path{doi:10.1093/bib/bbad093}}.

\bibitem{Liu2017}
Q.~Liu, J.~Wang, Y.~Zhao, et~al., {Identification of active miRNA promoters
  from nuclear run-on RNA sequencing}, Nucleic Acids Research 45~(13) (2017)
  e121--e121.
\newblock \href {https://doi.org/10.1093/nar/gkx318}
  {\path{doi:10.1093/nar/gkx318}}.

\bibitem{Rie2017}
D.~d.~Rie, I.~Abugessaisa, T.~Alam, et~al., {An integrated expression atlas of
  miRNAs and their promoters in human and mouse}, Nature Biotechnology 35
  (2017) 872–878.
\newblock \href {https://doi.org/10.1038/nbt.3947}
  {\path{doi:10.1038/nbt.3947}}.

\bibitem{Mai2022}
D.~H.~A. Mai, L.~T. Nguyen, E.~Y. Lee, Tssnote-cyaprombert: Development of an
  integrated platform for highly accurate promoter prediction and visualization
  of synechococcus sp. and synechocystis sp. through a state-of-the-art natural
  language processing model bert, Frontiers in Genetics 13 (2022).
\newblock \href {https://doi.org/10.3389/fgene.2022.1067562}
  {\path{doi:10.3389/fgene.2022.1067562}}.

\bibitem{Weizhi2022}
W.~An, Y.~Guo, Y.~Bian, et~al., Modna: Motif-oriented pre-training for dna
  language model, in: Proceedings of the 13th ACM International Conference on
  Bioinformatics, Computational Biology and Health Informatics, BCB '22, 2022.
\newblock \href {https://doi.org/10.1145/3535508.3545512}
  {\path{doi:10.1145/3535508.3545512}}.

\bibitem{Li2023}
J.~Li, Z.~Wu, W.~Lin, et~al., {iEnhancer-ELM: improve enhancer identification
  by extracting position-related multiscale contextual information based on
  enhancer language models}, Bioinformatics Advances 3~(1) (2023) vbad043.
\newblock \href {https://doi.org/10.1093/bioadv/vbad043}
  {\path{doi:10.1093/bioadv/vbad043}}.

\bibitem{LUO2023}
H.~Luo, Y.~Li, H.~Liu, et~al., Senet: A deep learning framework for
  discriminating super- and typical enhancers by sequence information,
  Computational Biology and Chemistry 105 (2023) 107905.
\newblock \href
  {https://doi.org/https://doi.org/10.1016/j.compbiolchem.2023.107905}
  {\path{doi:https://doi.org/10.1016/j.compbiolchem.2023.107905}}.

\bibitem{Ligeti2024}
B.~Ligeti, I.~Szepesi-Nagy, B.~Bodnár, et~al., {ProkBERT} family: genomic
  language models for microbiome applications, Frontiers in Microbiology 14
  (2024).
\newblock \href {https://doi.org/10.3389/fmicb.2023.1331233}
  {\path{doi:10.3389/fmicb.2023.1331233}}.

\bibitem{Lister2009}
R.~Lister, M.~Pelizzola, R.~H. Dowen, et~al., Human dna methylomes at base
  resolution show widespread epigenomic differences, Nature 462 (2009)
  315--322.
\newblock \href {https://doi.org/10.1038/nature08514}
  {\path{doi:10.1038/nature08514}}.

\bibitem{meissner2005}
A.~Meissner, A.~Gnirke, G.~W. Bell, et~al., Reduced representation bisulfite
  sequencing for comparative high-resolution dna methylation analysis, Nucleic
  acids research 33~(18) (2005) 5868--5877.
\newblock \href {https://doi.org/10.1093/nar/gki901}
  {\path{doi:10.1093/nar/gki901}}.

\bibitem{flusberg2010}
B.~A. Flusberg, D.~R. Webster, J.~H. Lee, et~al., Direct detection of dna
  methylation during single-molecule, real-time sequencing, Nature methods
  7~(6) (2010) 461--465.
\newblock \href {https://doi.org/doi.org/10.1038/nmeth.1459}
  {\path{doi:doi.org/10.1038/nmeth.1459}}.

\bibitem{landan2012}
G.~Landan, N.~M. Cohen, Z.~Mukamel, et~al., Epigenetic polymorphism and the
  stochastic formation of differentially methylated regions in normal and
  cancerous tissues, Nature genetics 44~(11) (2012) 1207--1214.
\newblock \href {https://doi.org/doi.org/10.1038/ng.2442}
  {\path{doi:doi.org/10.1038/ng.2442}}.

\bibitem{treangen2012}
T.~J. Treangen, S.~L. Salzberg, Repetitive dna and next-generation sequencing:
  computational challenges and solutions, Nature Reviews Genetics 13~(1) (2012)
  36--46.
\newblock \href {https://doi.org/10.1038/nrg3117} {\path{doi:10.1038/nrg3117}}.

\bibitem{Jin2022}
J.~Jin, Y.~Yu, R.~Wang, et~al., idna-abf: multi-scale deep biological language
  learning model for the interpretable prediction of dna methylations, Genome
  Biology 23~(219) (2022).
\newblock \href {https://doi.org/10.1186/s13059-022-02780-1}
  {\path{doi:10.1186/s13059-022-02780-1}}.

\bibitem{Ji2021dnabert}
Y.~Ji, Z.~Zhou, H.~Liu, et~al., {{DNABERT}: pre-trained Bidirectional Encoder
  Representations from Transformers model for DNA-language in genome},
  Bioinformatics 37~(15) (2021) 2112--2120.
\newblock \href {https://doi.org/10.1093/bioinformatics/btab083}
  {\path{doi:10.1093/bioinformatics/btab083}}.

\bibitem{Zeng2023}
W.~Zeng, A.~Gautam, D.~H. Huson, {MuLan-Methyl—multiple transformer-based
  language models for accurate DNA methylation prediction}, GigaScience 12
  (2023) giad054.
\newblock \href {https://doi.org/10.1093/gigascience/giad054}
  {\path{doi:10.1093/gigascience/giad054}}.

\bibitem{sanh2020distilbert}
V.~Sanh, L.~Debut, J.~Chaumond, et~al., Distil{BERT}, a distilled version of
  {BERT}: smaller, faster, cheaper and lighter (2020).
\newblock \href {http://arxiv.org/abs/1910.01108} {\path{arXiv:1910.01108}}.

\bibitem{clark2020electra}
K.~Clark, M.-T. Luong, Q.~V. Le, et~al., {ELECTRA}: Pre-training text encoders
  as discriminators rather than generators (2020).
\newblock \href {http://arxiv.org/abs/2003.10555} {\path{arXiv:2003.10555}}.

\bibitem{Yu1997}
Y.~T. Yu, M.~D. Shu, J.~A. Steitz, A new method for detecting sites of
  2'-o-methylation in rna molecules, RNA 3~(3) (1997) 324--31.

\bibitem{Dong2012}
Z.-W. Dong, P.~Shao, L.-T. Diao, et~al., Rtl-p: a sensitive approach for
  detecting sites of 2'-o-methylation in rna molecules, Nucleic Acids Research
  40~(20) (2012) e157.
\newblock \href {https://doi.org/10.1093/nar/gks698}
  {\path{doi:10.1093/nar/gks698}}.

\bibitem{CHEN2016}
W.~Chen, P.~Feng, H.~Tang, et~al., Identifying 2'-o-methylationation sites by
  integrating nucleotide chemical properties and nucleotide compositions,
  Genomics 107~(6) (2016) 255--258.
\newblock \href {https://doi.org/10.1016/j.ygeno.2016.05.003}
  {\path{doi:10.1016/j.ygeno.2016.05.003}}.

\bibitem{Yang2018}
H.~Yang, H.~Lv, H.~Ding, et~al., irna-2om: A sequence-based predictor for
  identifying 2'-o-methylation sites in homo sapiens, Journal of Computational
  Biology 25~(11) (2018) 1266--1277.
\newblock \href {https://doi.org/10.1089/cmb.2018.0004}
  {\path{doi:10.1089/cmb.2018.0004}}.

\bibitem{Ao2021}
C.~Ao, Q.~Zou, L.~Yu, {NmRF: identification of multispecies RNA
  2’-O-methylation modification sites from RNA sequences}, Briefings in
  Bioinformatics 23~(1) (2021) bbab480.
\newblock \href {https://doi.org/10.1093/bib/bbab480}
  {\path{doi:10.1093/bib/bbab480}}.

\bibitem{Li2021}
H.~Li, L.~Chen, Z.~Huang, et~al., Deepome: A web server for the prediction of
  2'-o-me sites based on the hybrid cnn and blstm architecture, Frontiers in
  Cell and Developmental Biology 9 (2021).
\newblock \href {https://doi.org/10.3389/fcell.2021.686894}
  {\path{doi:10.3389/fcell.2021.686894}}.

\bibitem{Soylu2023}
N.~N. Soylu, E.~Sefer, Bert2ome: Prediction of 2'-o-methylation modifications
  from rna sequence by transformer architecture based on bert, IEEE/ACM
  Transactions on Computational Biology and Bioinformatics 20~(3) (2023)
  2177--2189.
\newblock \href {https://doi.org/10.1109/TCBB.2023.3237769}
  {\path{doi:10.1109/TCBB.2023.3237769}}.

\bibitem{Stanojevic2024}
D.~Stanojević, Z.~Li, S.~Bakić, et~al., {Rockfish}: A transformer-based model
  for accurate 5-methylcytosine prediction from nanopore sequencing, Nature
  Communications 15~(5580) (2024).
\newblock \href {https://doi.org/10.1038/s41467-024-49847-0}
  {\path{doi:10.1038/s41467-024-49847-0}}.

\bibitem{Gwak2022}
H.-J. Gwak, M.~Rho, Transformer-based embedding applied to classify bacterial
  species using sequencing reads, in: 2022 IEEE International Conference on Big
  Data and Smart Computing (BigComp), 2022, pp. 374--377.
\newblock \href {https://doi.org/10.1109/BigComp54360.2022.00084}
  {\path{doi:10.1109/BigComp54360.2022.00084}}.

\bibitem{Rajkumar2022}
U.~Rajkumar, S.~Javadzadeh, M.~Bafna, et~al., Deepvifi: Detecting oncoviral
  infections in cancer genomes using transformers, in: Proceedings of the 13th
  ACM International Conference on Bioinformatics, Computational Biology and
  Health Informatics, BCB '22, 2022.
\newblock \href {https://doi.org/10.1145/3535508.3545551}
  {\path{doi:10.1145/3535508.3545551}}.

\bibitem{Gwak2022_1}
H.-J. Gwak, M.~Rho, {ViBE: a hierarchical BERT model to identify eukaryotic
  viruses using metagenome sequencing data}, Briefings in Bioinformatics 23~(4)
  (2022) bbac204.
\newblock \href {https://doi.org/10.1093/bib/bbac204}
  {\path{doi:10.1093/bib/bbac204}}.

\bibitem{SHANG2021}
J.~Shang, Y.~Sun, {CHEER}: {HierarCHical taxonomic classification for viral
  mEtagEnomic data via deep leaRning}, Methods 189 (2021) 95--103.
\newblock \href {https://doi.org/10.1016/j.ymeth.2020.05.018}
  {\path{doi:10.1016/j.ymeth.2020.05.018}}.

\bibitem{ren2020}
J.~Ren, K.~Song, C.~Deng, et~al., Identifying viruses from metagenomic data
  using deep learning, Quantitative Biology 8 (2020) 64--77.
\newblock \href {https://doi.org/10.1007/s40484-019-0187-4}
  {\path{doi:10.1007/s40484-019-0187-4}}.

\bibitem{Tang2023}
X.~Tang, J.~Shang, Y.~Ji, et~al., {PLASMe: a tool to identify PLASMid contigs
  from short-read assemblies using transformer}, Nucleic Acids Research (2023)
  gkad578\href {https://doi.org/10.1093/nar/gkad578}
  {\path{doi:10.1093/nar/gkad578}}.

\bibitem{Wichmann2023}
A.~Wichmann, E.~Buschong, A.~Müller, et~al., {MetaTransformer: deep
  metagenomic sequencing read classification using self-attention models}, NAR
  Genomics and Bioinformatics 5~(3) (2023) lqad082.
\newblock \href {https://doi.org/10.1093/nargab/lqad082}
  {\path{doi:10.1093/nargab/lqad082}}.

\bibitem{Liangdeepmicrobes2020}
Q.~Liang, P.~W. Bible, Y.~Liu, et~al., {DeepMicrobes: taxonomic classification
  for metagenomics with deep learning}, NAR Genomics and Bioinformatics 2~(1)
  (2020) lqaa009.
\newblock \href {https://doi.org/10.1093/nargab/lqaa009}
  {\path{doi:10.1093/nargab/lqaa009}}.

\bibitem{Yamada2022}
K.~Yamada, M.~Hamada, {Prediction of RNA–protein interactions using a
  nucleotide language model}, Bioinformatics Advances 2~(1) (2022) vbac023.
\newblock \href {https://doi.org/10.1093/bioadv/vbac023}
  {\path{doi:10.1093/bioadv/vbac023}}.

\bibitem{HLuo2023}
H.~Luo, W.~Shan, C.~Chen, et~al., {Improving language model of human genome for
  DNA–protein binding prediction based on task-specific pre-training},
  Interdisciplinary Sciences: Computational Life Sciences 15 (2023) 32--43.
\newblock \href {https://doi.org/10.1007/s12539-022-00537-9}
  {\path{doi:10.1007/s12539-022-00537-9}}.

\bibitem{Wang2023_1}
X.~Wang, M.~Zhang, C.~Long, et~al., Self-attention based neural network for
  predicting rna-protein binding sites, IEEE/ACM Transactions on Computational
  Biology and Bioinformatics 20~(2) (2023) 1469--1479.
\newblock \href {https://doi.org/10.1109/TCBB.2022.3204661}
  {\path{doi:10.1109/TCBB.2022.3204661}}.

\bibitem{Wu2021}
K.~Wu, K.~E. Yost, B.~Daniel, et~al., Tcr-bert: learning the grammar of t-cell
  receptors for flexible antigen-xbinding analyses, bioRxiv (2021).
\newblock \href {https://doi.org/10.1101/2021.11.18.469186}
  {\path{doi:10.1101/2021.11.18.469186}}.

\bibitem{Zhang2022}
Y.~Zhang, Y.~Liu, Z.~Wang, et~al., Uncovering the relationship between
  tissue-specific tf-dna binding and chromatin features through a
  transformer-based model, Genes 13~(11) (2022).
\newblock \href {https://doi.org/10.3390/genes13111952}
  {\path{doi:10.3390/genes13111952}}.

\bibitem{Ding2023}
P.~Ding, Y.~Wang, X.~Zhang, et~al., {{DeepSTF}: predicting transcription factor
  binding sites by interpretable deep neural networks combining sequence and
  shape}, Briefings in Bioinformatics 24~(4) (2023) bbad231.
\newblock \href {https://doi.org/10.1093/bib/bbad231}
  {\path{doi:10.1093/bib/bbad231}}.

\bibitem{Ghosh2024}
N.~Ghosh, D.~Santoni, I.~Saha, et~al., Predicting transcription factor binding
  sites with deep learning, International Journal of Molecular Sciences 25~(9)
  (2024).
\newblock \href {https://doi.org/10.3390/ijms25094990}
  {\path{doi:10.3390/ijms25094990}}.

\bibitem{zhang2023dnagpt}
D.~Zhang, W.~Zhang, B.~He, et~al., Dnagpt: A generalized pretrained tool for
  multiple dna sequence analysis tasks (2023).
\newblock \href {http://arxiv.org/abs/2307.05628} {\path{arXiv:2307.05628}}.

\bibitem{Fishman2023}
V.~Fishman, Y.~Kuratov, M.~Petrov, et~al., Gena-lm: A family of open-source
  foundational models for long dna sequences, bioRxiv (2023).
\newblock \href {https://doi.org/10.1101/2023.06.12.544594}
  {\path{doi:10.1101/2023.06.12.544594}}.

\bibitem{Dalla-Torre2023}
H.~Dalla-Torre, L.~Gonzalez, J.~M. Revilla, et~al., The nucleotide transformer:
  Building and evaluating robust foundation models for human genomics, bioRxiv
  (2023).
\newblock \href {https://doi.org/10.1101/2023.01.11.523679}
  {\path{doi:10.1101/2023.01.11.523679}}.

\bibitem{zhou2023}
Z.~Zhou, Y.~Ji, W.~Li, et~al., Dnabert-2: Efficient foundation model and
  benchmark for multi-species genome (2023).
\newblock \href {http://arxiv.org/abs/2306.15006} {\path{arXiv:2306.15006}}.

\bibitem{Clauwaert2023}
J.~Clauwaert, Z.~McVey, R.~Gupta, et~al., {TIS Transformer: remapping the human
  proteome using deep learning}, NAR Genomics and Bioinformatics 5~(1) (2023)
  lqad021.
\newblock \href {https://doi.org/10.1093/nargab/lqad021}
  {\path{doi:10.1093/nargab/lqad021}}.

\bibitem{du2023}
Y.~Du, Z.~Li, Q.~He, et~al., Dpcipi: A pre-trained deep learning model for
  estimation of cross-immunity between drifted strains of influenza a/h3n2
  (2023).
\newblock \href {http://arxiv.org/abs/2302.00926} {\path{arXiv:2302.00926}}.

\bibitem{Bai2022}
Z.~Bai, Y.~z.~Zhang, S.~Miyano, et~al., {Identification of bacteriophage genome
  sequences with representation learning}, Bioinformatics 38~(18) (2022)
  4264--4270.
\newblock \href {https://doi.org/10.1093/bioinformatics/btac509}
  {\path{doi:10.1093/bioinformatics/btac509}}.

\bibitem{Kieft2020}
K.~Kieft, Z.~Zhou, K.~Anantharaman, {{VIBRANT}: automated recovery, annotation
  and curation of microbial viruses, and evaluation of viral community function
  from genomic sequences}, Microbiome 8~(90) (02 2020).
\newblock \href {https://doi.org/10.1186/s40168-020-00867-0}
  {\path{doi:10.1186/s40168-020-00867-0}}.

\bibitem{Guo2021_sorter}
J.~Guo, B.~Bolduc, A.~A. Zayed, et~al., {{VirSorter2}: a multi-classifier,
  expert-guided approach to detect diverse {DNA} and {RNA} viruses}, Microbiome
  9~(37) (07 2021).
\newblock \href {https://doi.org/10.1186/s40168-020-00990-y}
  {\path{doi:10.1186/s40168-020-00990-y}}.

\bibitem{Auslander2020}
N.~Auslander, A.~B. Gussow, S.~Benler, et~al., {Seeker: alignment-free
  identification of bacteriophage genomes by deep learning}, Nucleic Acids
  Research 48~(21) (2020) e121--e121.
\newblock \href {https://doi.org/10.1093/nar/gkaa856}
  {\path{doi:10.1093/nar/gkaa856}}.

\bibitem{Lee2022}
D.~Lee, J.~Yang, S.~Kim, {Learning the histone codes with large genomic windows
  and three-dimensional chromatin interactions using transformer}, Nature
  Communications 13~(6678) (2022).
\newblock \href {https://doi.org/10.1038/s41467-022-34152-5}
  {\path{doi:10.1038/s41467-022-34152-5}}.

\bibitem{Raad2021}
J.~Raad, L.~A. Bugnon, D.~H. Milone, et~al., {{miRe2e}: a full end-to-end deep
  model based on transformers for prediction of pre-miRNAs}, Bioinformatics
  38~(5) (2021) 1191--1197.
\newblock \href {https://doi.org/10.1093/bioinformatics/btab823}
  {\path{doi:10.1093/bioinformatics/btab823}}.

\bibitem{Zhang2023}
Y.~Zhang, F.~Ge, F.~Li, et~al., Prediction of multiple types of rna
  modifications via biological language model, IEEE/ACM Transactions on
  Computational Biology and Bioinformatics (2023) 1--11\href
  {https://doi.org/10.1109/TCBB.2023.3283985}
  {\path{doi:10.1109/TCBB.2023.3283985}}.

\bibitem{Jurenaite2022}
N.~Jurenaite, D.~Leon-Perinan, V.~Donath, et~al., Setquence \& setomic: Deep
  set transformer-based representations of cancer multi-omics, in: 2022 IEEE
  Conference on Computational Intelligence in Bioinformatics and Computational
  Biology (CIBCB), 2022, pp. 1--9.
\newblock \href {https://doi.org/10.1109/CIBCB55180.2022.9863058}
  {\path{doi:10.1109/CIBCB55180.2022.9863058}}.

\bibitem{Avsec2021}
Z.~Avsec, V.~Agarwal, D.~Visentin, et~al., Effective gene expression prediction
  from sequence by integrating long-range interactions, Nature Methods 18
  (2021) 1196–1203.
\newblock \href {https://doi.org/10.1038/s41592-021-01252-x}
  {\path{doi:10.1038/s41592-021-01252-x}}.

\bibitem{Kelley2020}
D.~R. Kelley, Cross-species regulatory sequence activity prediction, PLOS
  Computational Biology 16~(7) (2020) 1--27.
\newblock \href {https://doi.org/10.1371/journal.pcbi.1008050}
  {\path{doi:10.1371/journal.pcbi.1008050}}.

\bibitem{Nguyen2023}
E.~Nguyen, M.~Poli, M.~Faizi, et~al., {HyenaDNA}: Long-range genomic sequence
  modeling at single nucleotide resolution, in: Advances in Neural Information
  Processing Systems, Vol.~36, 2023, pp. 43177--43201.

\bibitem{zaheer2021big}
M.~Zaheer, G.~Guruganesh, A.~Dubey, et~al., Big bird: Transformers for longer
  sequences (2021).
\newblock \href {http://arxiv.org/abs/2007.14062} {\path{arXiv:2007.14062}}.

\bibitem{choromanski2022}
K.~Choromanski, V.~Likhosherstov, D.~Dohan, et~al., Rethinking attention with
  performers (2022).
\newblock \href {http://arxiv.org/abs/2009.14794} {\path{arXiv:2009.14794}}.

\bibitem{xie2023chunk}
J.~Xie, P.~Cheng, X.~Liang, et~al., Chunk, align, select: A simple
  long-sequence processing method for transformers (2023).
\newblock \href {http://arxiv.org/abs/2308.13191} {\path{arXiv:2308.13191}}.

\bibitem{sennrich2016}
R.~Sennrich, B.~Haddow, A.~Birch, Neural machine translation of rare words with
  subword units, in: Proceedings of the 54th Annual Meeting of the Association
  for Computational Linguistics (Volume 1: Long Papers), 2016, pp. 1715--1725.
\newblock \href {https://doi.org/10.18653/v1/P16-1162}
  {\path{doi:10.18653/v1/P16-1162}}.

\end{thebibliography}

\end{document}